\documentclass{article}

\usepackage{microtype}
\usepackage{graphicx}
\usepackage{subfigure}
\usepackage{booktabs} 

\usepackage{amsmath}
\usepackage{etoolbox}

\usepackage{amssymb}
\usepackage{algorithmic}
\usepackage{forloop}
\usepackage{fancyhdr}
\usepackage{stmaryrd}
\usepackage{textcomp}
\usepackage{bm}

\usepackage{amssymb}
\usepackage{booktabs}
\usepackage{multirow}
\usepackage{kotex}
\usepackage{tablefootnote}
\usepackage{footnote}
\usepackage[dvipsnames,table,xcdraw]{xcolor}
\usepackage{lineno}
\usepackage{hyperref}
\usepackage{graphicx}
\usepackage{ulem} 
\usepackage{textcomp}
\usepackage{latexsym}
\usepackage{graphicx}
\usepackage{tabularx}
\usepackage{graphics}
\usepackage{adjustbox}
\usepackage{algorithmic}
\graphicspath{{./figures/}}

\usepackage{relsize}
\usepackage{mathtools}
\usepackage{hyperref}

\usepackage[accepted]{icml2021}

\icmltitlerunning{AnoViT: Unsupervised Anomaly Detection and Localization with Vision Transformer-based Encoder-Decoder}

\begin{document}

\twocolumn[
\icmltitle{AnoViT: Unsupervised Anomaly Detection and Localization with Vision Transformer-based Encoder-Decoder}



\icmlsetsymbol{equal}{*}

\begin{icmlauthorlist}
\icmlauthor{Yunseung Lee}{Korea}
\icmlauthor{Pilsung Kang}{Korea}
\end{icmlauthorlist}

\icmlaffiliation{Korea}{School of Industrial Management \& Engineering,
College of Engineering,
Korea University, Seoul, Korea}

\icmlcorrespondingauthor{Pilsung Kang}{pilsung\_kang@korea.ac.kr}

\icmlkeywords{Anomaly Detection, Anomaly Localization, Vision Transformer, MVTecAD}

\vskip 0.3in ]



\printAffiliationsAndNotice{}  

\begin{abstract}
Image anomaly detection problems aim to determine whether an image is abnormal, and to detect anomalous areas. These methods are actively used in various fields such as manufacturing, medical care, and intelligent information. Encoder-decoder structures have been widely used in the field of anomaly detection because they can easily learn normal patterns in an unsupervised learning environment and calculate a score to identify abnormalities through a reconstruction error indicating the difference between input and reconstructed images. Therefore, current image anomaly detection methods have commonly used convolutional encoder-decoders to extract normal information through the local features of images. However, they are limited in that only local features of the image can be utilized when constructing a normal representation owing to the characteristics of convolution operations using a filter of fixed size. Therefore, we propose a vision transformer-based encoder-decoder model, named AnoViT, designed to reflect normal information by additionally learning the global relationship between image patches, which is capable of both image anomaly detection and localization. The proposed approach constructs a feature map that maintains the existing location information of individual patches by using the embeddings of all patches passed through multiple self-attention layers. The proposed AnoViT model performed better than the convolution-based model on three benchmark datasets. In MVTecAD, which is a representative benchmark dataset for anomaly localization, it showed improved results on 10 out of 15 classes compared with the baseline. Furthermore, the proposed method showed good performance regardless of the class and type of the anomalous area when localization results were evaluated qualitatively.
\end{abstract}

\section{Introduction}
\label{sec:intro}
Image anomaly detection methods use image data to detect data samples with a distribution that differs substantially from that of normal images. Image anomaly localization aims to find the location of defects in a given image, and is referred to as anomaly detection from a pixel perspective \cite{yang2021visual}. Anomaly detection in image data is an essential technology in various fields such as manufacturing, medical care, and intelligent information \cite{mohammadi2021image}. For example, image anomaly detection is used in the manufacturing field to make quick decisions during visual inspection to identify defects in manufactured products \cite{yang2021visual}. In the medical field, it is used to detect tumors based on images collected from fMRI and CT \cite{guo2018medical}. Moreover, image anomaly detection is used in the intelligent information field to detect abnormal behaviors in CCTV videos \cite{sultani2018real}. By providing information on whether an image contains defects as well as on their locations, image anomaly detection can help experts make quick decisions and hence contributes to improved work efficiency \cite{nakazawa2019anomaly}. Therefore, the development of advanced image anomaly detection and localization methods is of considerable interest.

As research on deep learning has been actively conducted, many deep learning-based image anomaly detection methods have been proposed. Image anomaly detection and localization methods using deep learning can be divided into encoder-decoder-based methods and generation-based methods. Encoder-decoder-based methods train a model by compressing a normal input image and reconstructing it to be similar to the original image \cite{bergmann2019mvtec, DBLP:conf/visapp/BergmannLFSS19, chow2020anomaly}. Previous studies have generally used a convolutional autoencoder (CAE) structure, and some of these works adopted a U-Net-based model. Anomaly detection is performed using the difference between the original image and the image reconstructed through the encoder-decoder. Generation-based methods learn a distribution of normal data by generating images with a distribution similar to that of an original image \cite{schlegl2017unsupervised, schlegl2019f, venkataramanan2020attention, dai2019diagnosing}. Anomaly detection methods may be broadly  classified into generative adversarial network (GAN)- and variational autoencoder (VAE)-based models depending on the learning method adopted. Both models identify anomalies based on the difference between an original image and an image generated from a learned distribution.

Recent image anomaly detection methodologies can be classified according to their input image processing methods. One category uses the whole image, whereas the other divides an image into patches. Most of the previous studies used whole images, but patch-level images have been widely used in more recent works \cite{yi2020patch, defard2021padim, wang2021glancing}. Patch-level input is used to generate image representation vectors that contain rich information. If segmented image patches are used, embeddings can be generated per patch. Hence, information for each detailed area in the image can be reflected well \cite{defard2021padim}. Furthermore, patch-level input simplifies the task of relative location prediction, which is a pretext task for self-supervised learning methods \cite{doersch2015unsupervised, yi2020patch}.

In terms of model structure, existing deep learning-based image anomaly detection methods generally learn an image representation using a convolutional neural network (CNN)-type model. However, because CNNs only learn local information, they have difficulty learning the global context of an image \cite{raghu2021vision}. The recently published vision transformer (ViT) was designed to additionally learn the global context of an image by using a patch-level image and attention operation. Hence, ViT alleviates this problem associated with CNNs \cite{dosovitskiy2020image}. 

Given this background, we propose an AnoViT model capable of both image anomaly detection and localization using the ViT-based encoder-decoder structure and reconstruction error. The proposed method generates image embeddings containing rich information for each area by processing patch-level images. In addition, it also utilizes the global information by learning the relationship between patches through an attention operation. The proposed method was applied to the MVTecAD dataset, which is most widely used in the image anomaly detection field. The results confirmed that the proposed method achieved improved image anomaly detection and localization performance by utilizing both the local and global information.

The remainder of this paper is organized as follows. Section 2 introduces prior research on the encoder-decoder-based model and the structure of the ViT model. Section 3 introduces the proposed method, and Section 4 covers the quantitative and qualitative experimental results. Finally, Section 5 summarizes and concludes the work, and provides some possible directions for future research.

\section{Related Work}

\subsection{Encoder-Decoder-based Image Anomaly Detection}
The encoder-decoder-based method learns encoder $\mathcal{E}$ and decoder $\mathcal{D}$ by minimizing a reconstruction error $\mathcal{L}$ between the original image $\mathcal{X_N}$ and an image $\mathcal{D(E(X_N))}$ reconstructed from the normal image $\mathcal{X_N}$ compressed with encoder $\mathcal{E}$.

\begin{equation}\mathcal{L}=\frac{1}{m}\sum \left\|\mathcal{X_N^\mathnormal{i}-D(E(X_N^\mathnormal{i}))}\right\|^2 ,\label{eq:1}\end{equation}

As shown in \eqref{eq:1}, the neural network of the encoder-decoder structure learns the distribution $\mathcal{\mathit{p}_N}$ of the normal image inherently in the optimization process. Because $\mathcal{D(E(X_N))}$ has been trained to reconstruct the normal data well, the reconstruction error increases when abnormal data are provided as input. Based on this principle, the encoder-decoder-based method determines an image to be abnormal if the reconstruction error is higher than a certain threshold. Conversely, images are identified as normal if the reconstruction error is lower than a certain threshold \cite{mohammadi2021image}.

A convolutional autoencoder (CAE) is a representative model used in encoder-decoder-based image anomaly detection methods. CAE models are suitable for extracting image representations because each layer of the CAE model uses a convolution operation \cite{dosovitskiy2014discriminative}. Moreover, CAE models have the advantage of being able to identify anomalies at the image level and the pixel level by using a score map that differs between the original image and the reconstructed image \cite{sabokrou2018deep}. CAE models are used in image anomaly detection and localization in various fields for this reason. Specifically, they have performed well in defect detection in semiconductor wafer bin maps \cite{nakazawa2019anomaly}, detection of cracks in images of concrete structures \cite{chow2020anomaly}, detection of defects in images of fabric texture \cite{mei2018unsupervised, DBLP:conf/visapp/BergmannLFSS19}, and detection of abnormal behaviors in videos \cite{zhao2017spatio}, among other applications.

In CAE-based image anomaly detection, the difference between the original and reconstructed images/videos is typically learned by calculating $l_1$-loss or $l_2$-loss \cite{nakazawa2019anomaly, chow2020anomaly, mei2018unsupervised}. When detecting defects in images of a given texture, a structured similarity index (SSIM) is sometimes added as a loss function. This technique helps to reflect the similarity between images and is especially effective for black-and-white images \cite{DBLP:conf/visapp/BergmannLFSS19}. In addition, multi-scale encoding and denoised CAE are used \cite{mei2018unsupervised}, or a latent vector is extracted by adding a memory module \cite{gong2019memorizing} to construct a representation that reflects features of the normal data well. Moreover, the performance of anomaly detection and localization on the publicly available MVTec benchmark dataset was compared with other methodologies such as SSIM-CAE, AnoGAN, and CNN Feature Dictionary\cite{bergmann2019mvtec}, and the experimental results verified that the $l_2$-CAE exhibited the best performance.

\subsection{Vision Transformer}
Transformer was originally designed as a sequence-to-sequence language model with self-attention mechanisms based on encoder-decoder structure to solve natural language processing (NLP) tasks. By using the attention mechanism, the transformers have solved the problem of previous sequence-to-sequence models, which were not able adequately to learn the relationship between distant words in a sentence. Consequently, transformers have significantly improved performance in various NLP tasks, such as machine translation and question and answer \cite{vaswani2017attention}. Recently, \cite{dosovitskiy2020image} first introduced Vision Transformers (ViTs), which were developed by modifying the encoder of a transformer model to perform image classification tasks. Since then, ViTs have been used in various vision fields, such as object detection and semantic segmentation \cite{carion2020end, chen2021transunet, liu2021swin}.

\begin{figure}[htp]
\centering
\includegraphics[scale=0.14]{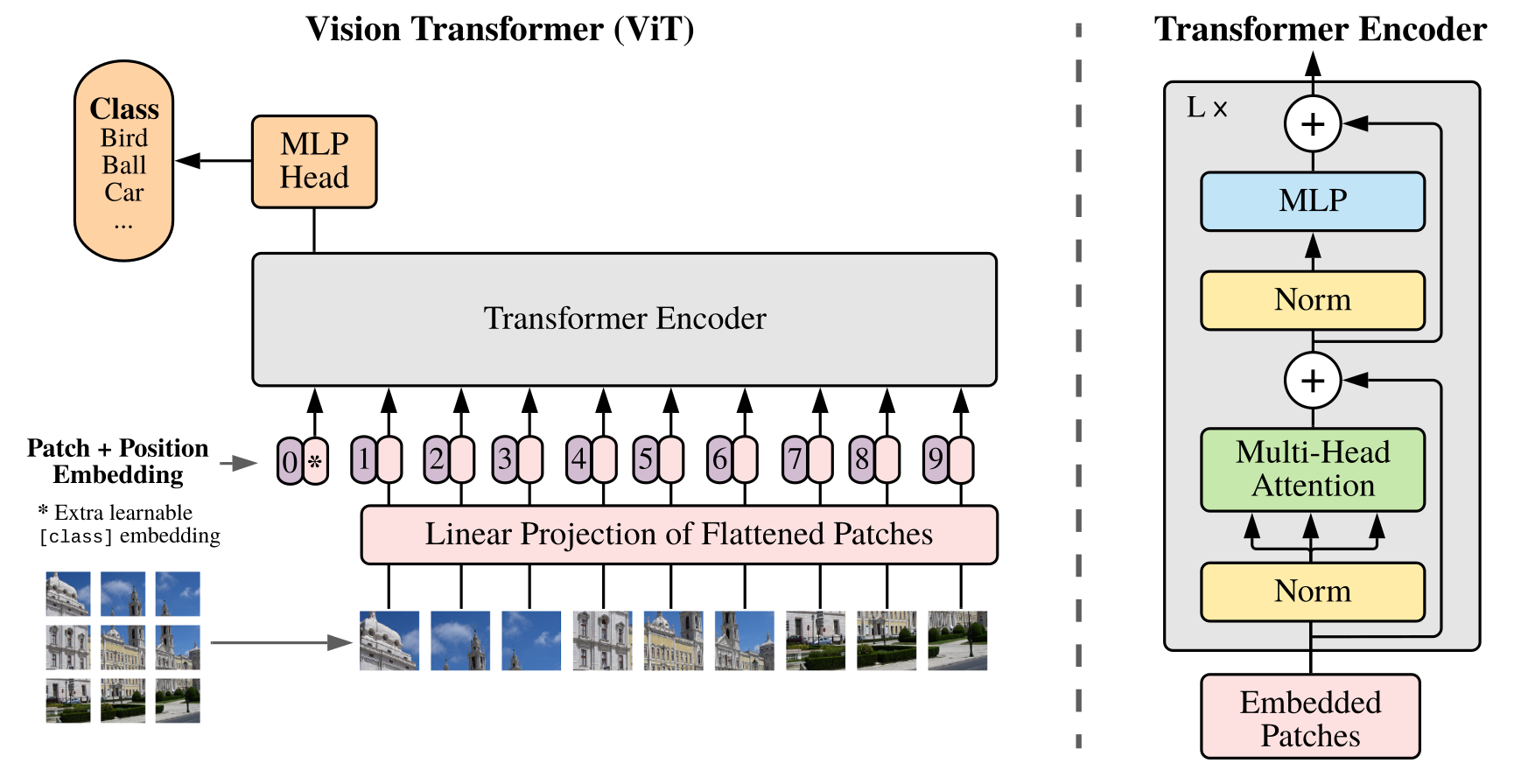}
\caption{Architecture of original ViT}
\label{fig1}
\end{figure}

As shown in Fig. \ref{fig1}, ViT uses an image segmented into patch units as input. The patches are transformed into a two-dimensional sequence to learn the relationship between each patch through multi-head self-attention. Considering the operation of ViT models in detail, the image $\mathcal{X}\in\mathbb{R}^{H{\times}W{\times}C}$ is reshaped into patches $x_p\in\mathbb{R}^{N{\times}(P^2\cdot{C})}$ and then mapped to $D$ dimensions (image size: $(H,W,C)$, patch size: $(P,P)$, $N=HW/P^2$). After these patches pass through a learnable linear projection, two-dimensional patch embeddings are derived as an output. The positional embedding $\pmb{\mathrm{E}}_{pos}\in\mathbb{R}^{(N+1){\times}D}$ is added to the patch embedding $\pmb{\mathrm{E}}$, which concatenates the $[cls]$ token $z_0^0=x_{cls}$, to preserve position information. The embeddings pass through layers composed of multi-head self-attention, an MLP block, and Layer Normalization($LN$) by the number of blocks. Among the patch embeddings derived from the transformer encoder, only the $[cls]$ token is used as an input to the MLP head to perform the image classification task.

As benchmark datasets for image classification tasks, CIFAR-10, CIFAR-100, and ImageNet were used to conduct an experiment on a ViT and CNN-based ResNet pretrained on JFT-300M. The experimental results demonstrated that ViT-type models outperformed the CNN models \cite{dosovitskiy2020image}.

\begin{figure*}[t!]
\begin{center}
\includegraphics[scale=0.45]{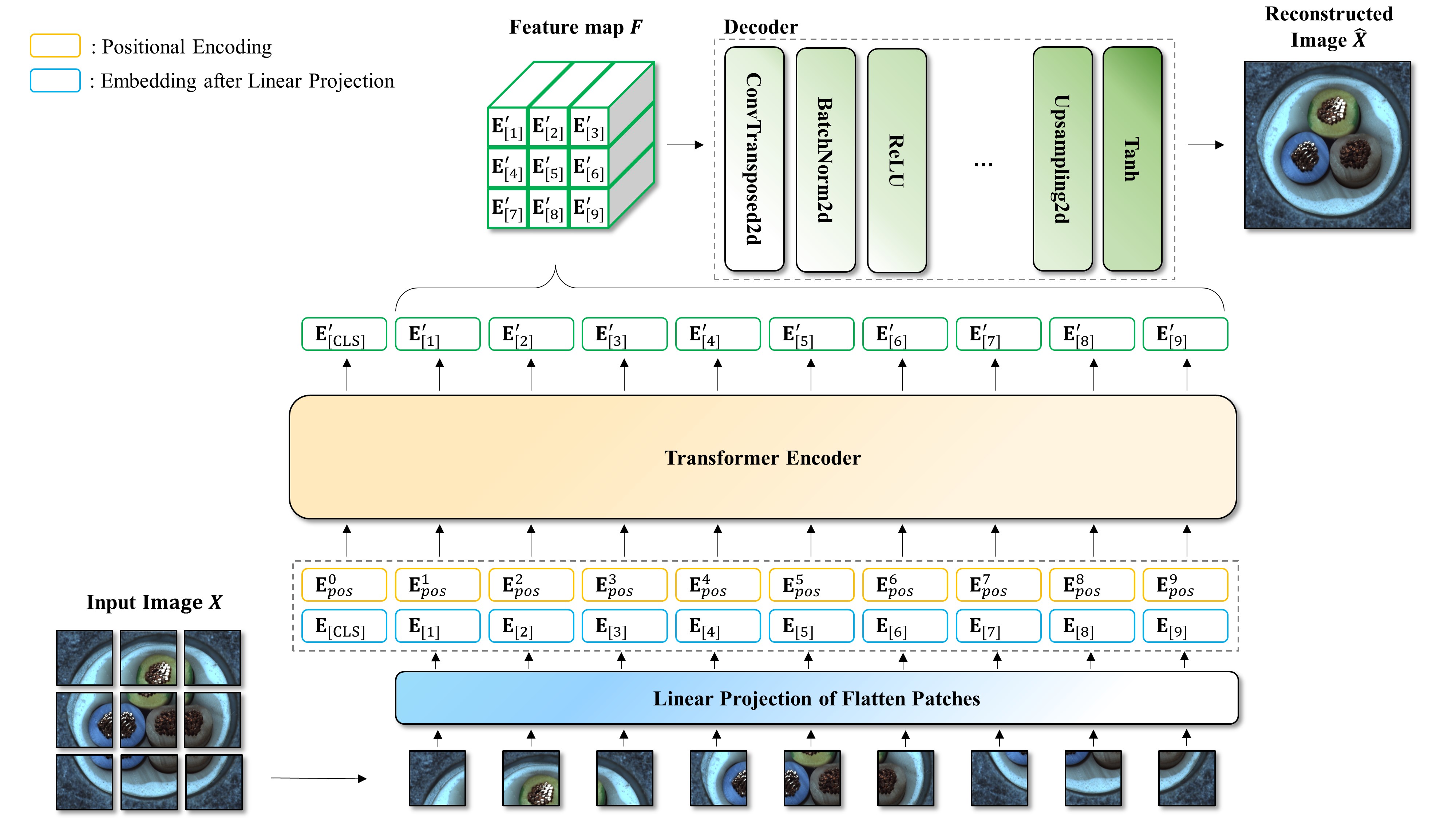}
\end{center}
\caption{Architecture of AnoViT}
\label{fig2}
\end{figure*}

\section{Proposed Method}
\subsection{ViT-based Encoder-Decoder}
In this study, we propose an encoder-decoder-based method—which consists of an encoder using ViT and a decoder containing a convolutional layer—and a model structure that can perform both image anomaly detection and localization with a reconstruction error. The proposed method, AnoViT,  reflects the relationships between image patches in the representation through the ViT's multi-head self-attention (MSA). Furthermore, the proposed method minimizes the reconstruction error, the difference between the input image and the reconstructed image, and learns the distribution of the normal data. The structure of AnoViT is shown in Fig. \ref{fig2}.

The encoder is processed in the same manner as in existing ViT, and provided as an input to the model. The two-dimensional embedded patches $\pmb{\mathrm{E}} \in \mathbb{R}^{(N+1){\times}D}$ learn the relationships between each patch through $k$ self-attention (SA)--namely, multi-head self-attention (MSA)--operations according to the equations given below. $\mathcal{A}$ denotes the attention weight matrix, and $\mathcal{A}_{ij}$ indicates the pairwise similarity of query $\pmb{\mathrm{q}}^i$ and key $\pmb{\mathrm{k}}^j$. Moreover, the attention information is reflected in the existing embeddings through the weighted sum of all elements value $\pmb{\mathrm{v}}$ of patch embedding $\pmb{\mathrm{E}}$. 
\begin{gather}
[\ \pmb{\mathrm{q}},\pmb{\mathrm{k}},\pmb{\mathrm{v}} ]\ = \pmb{\mathrm{E}}\pmb{\mathrm{U}}_{qkv}, \qquad \pmb{\mathrm{U}}_{qkv} \in\mathbb{R}^{D{\times}3D_h},
\label{eq:2} \\
\mathcal{A}=\mathrm{softmax}(\pmb{\mathrm{q}}\pmb{\mathrm{k}}^\top/\sqrt{D_h}), \quad \mathcal{A} \in \mathbb{R}^{(N+1){\times}(N+1)},
\label{eq:3}
\end{gather}
\begin{equation}
\mathrm{SA}(\pmb{\mathrm{E}}) = \mathcal{A}\pmb{\mathrm{v}}, 
\label{eq:4}
\end{equation}
\begin{equation} \label{eq:5}
\begin{split}
\mathrm{MSA}(\pmb{\mathrm{E}}) = [\ \mathrm{SA}_1(\pmb{\mathrm{E}}); \mathrm{SA}_2(\pmb{\mathrm{E}}); ... ;\mathrm{SA}_k(\pmb{\mathrm{E}})] \pmb{\mathrm{U}}_{msa}, \\ \pmb{\mathrm{U}}_{msa} \in \mathbb{R}^{k\cdot{D_h}{\times}D}.
\end{split}
\end{equation}

The adjacent patch information is used to construct embeddings for a single patch, and information is received from all patches in the image. Hence, rich information can be contained in embeddings. That is, patch embeddings can be extracted that reflect the global context of the image by using the attention-based ViT encoder.

In contrast to existing ViTs that derive the output using the $[cls]$ token and an additional MLP head, the proposed AnoViT model uses embeddings of each image patch from the output of the ViT. In anomaly detection and localization, it is crucial to calculate the reconstruction error at the image and pixel levels. Hence, a representation containing rich information on the normal image is required to calculate the reconstruction error in the image reconstruction process. Thus, the encoder is configured to extract patch embeddings that contain detailed information for each area of the image patch.

The $[cls]$ token of the patch embeddings $\pmb{\mathrm{E}}^{'} \in \mathbb{R}^{(N+1){\times}D}$ calculated through the decoder ViT encoder is excluded, and the remaining embeddings are rearranged to match the existing positions in the patch image and used as the feature map $\mathcal{F}\in\mathbb{R}^{N^{\frac{1}{2}}{\times}N^{\frac{1}{2}}{\times}D}$. Several CAE models tend to extract a latent vector through the encoder and then pass it through the fully-connected layer. However, the proposed method rearranges the feature map $\mathcal{F}$ in three dimensions. Therefore, the feature map $\mathcal{F}$ can be directly used as an input to the decoder without being passed through an additional layer. Also, because the fully-connected layer is not used, the feature map $\mathcal{F}$ can preserve the spatial information between patches well. In the proposed method, the decoder is configured to reconstruct an image that is the same size as the original image. Precisely, the decoder reconstructs an image $\hat{\mathcal{X}}\in\mathbb{R}^{H{\times}W{\times}C}$ from the feature map $\mathcal{F}$ based on transposed convolution layers.

\subsection{Anomaly Detection and Localization}
The proposed approach uses reconstruction error to identify anomalies. The $l_2$-distance between the original image and the reconstructed image derived from the output of the ViT-based encoder-decoder model $\mathit{f}$ is calculated for each pixel, and the score map $\mathcal{M}$ is calculated by taking the average pooling across the channels. In addition, $X_{ij}$ in \eqref{eq:6} denotes the $(i,j)$-th pixel in the input image.
\begin{gather}
\mathcal{M} = \left\|\mathcal{X}_{ij}-\mathit{f}(\mathcal{X}_{ij})\right\|_2, \label{eq:6}
\\
\mathit{s_a} = \max{\mathcal{M}}. \label{eq:7}
\end{gather}
Because the proposed method learns the distribution of the normal image, the reconstruction error increases when an abnormal image is provided as an input because the areas containing anomalies are not reconstructed well. The proposed method utilizes this fact to detect anomalies.Anomaly score $\mathit{s_a}$ is calculated by \eqref{eq:7}, which takes the maximum value in the score map $\mathcal{M}$. It is then used to identify the presence of an anomaly in anomaly detection. In the case of anomaly localization, score map $\mathcal{M}$ is used to determine whether each pixel in the image is abnormal or not.

\section{Experimental Setup}
\subsection{Data}
The MNIST, CIFAR10, and MVTecAD datasets were used to verify the anomaly detection and localization performance of the proposed method. The anomaly detection performance was evaluated using the MNIST, CIFAR10, and MVTecAD datasets, and the anomaly localization performance was verified using the MVTecAD dataset.

\subsubsection{MNIST, CIFAR10}
Both the MNIST and CIFAR10 datasets comprise image data composed of ten classes. They both consist of normal and abnormal datasets in the same way as the general experimental settings in one-class classification. Images of one class were considered to be normal and used as training data, whereas the remaining nine classes were defined as abnormal. The test dataset consisted of images of both the normal and abnormal classes. Images for both datasets were resized to (224, 224, 3) and used to perform training and evaluation.

The MNIST dataset comprises digits ranging from 0 to 9. It includes approximately 6,000 images per class for the training data. 80\% of the data were used to perform training and to verify the model's performance in the training process. The testing data consist of 10,000 images, including both normal and abnormal classes. 

The CIFAR10 dataset includes image data on ten objects, with 5,000 images per class for the training data. Four thousand, five hundred images were used to train the model, and the remaining 500 images were used to verify its performance. The testing data comprised 10,000 images, including both normal and abnormal classes.

\begin{table}
\caption{Overall performance of anomaly detection and localization.}
\label{tab:overall result}
\centering 
\renewcommand{\arraystretch}{1.0}
\setlength{\tabcolsep}{5pt}
\begin{tabular}{llccc}
\toprule
\multirow{2}{*}{\textbf{Dataset (Task)}} & \multirow{2}{*}{$l_2$\textbf{-CAE}}   &  \textbf{Ours} &  \textbf{Relative}  \\ 
& &  \textbf{(AnoViT)} & \textbf{Improvement} \\

\midrule

 \multirow{1}{*}{MNIST (Det)} & 0.908 & \uline{\textbf{0.924}} & \uline{\textbf{1.774\%}} \\
 \multirow{1}{*}{CIFAR10 (Det)} & 0.569 & \uline{\textbf{0.601}} & \uline{\textbf{5.624\%}} \\
 \multirow{1}{*}{MVTecAD (Det)} & 0.730  & \uline{\textbf{0.780}}  & \uline{\textbf{6.850\%}} \\ 
 \multirow{1}{*}{MVTecAD (Loc)} & 0.810  & \uline{\textbf{0.830}}  & \uline{\textbf{2.470\%}} \\ 

\bottomrule
\end{tabular}
\end{table}

\subsubsection{MVTec Anomaly Detection Dataset}
The MVTecAD dataset is an image data on 15 products. The product group is largely classified into object and texture classes. The object class contains ten products, and the texture class contains five. The MVTecAD dataset consists of 3,629 normal data and 1,725 abnormal data, and the images were resized to (384, 384, 3) for training and testing. Because normal and abnormal images per product class are included in the dataset, the model was trained and evaluated its anomaly detection performance for each product class. The MVTecAD dataset provides image-level labels along with the pixel-level ground-truth for the location of anomalies. The MVTecAD dataset was also used to evaluate the anomaly localization performance based on this information.

\subsection{Baselines and Evaluation Metrics}
Because the proposed method is a ViT-based encoder-decoder model deisgned to perform anomaly detection using the reconstruction error, the $l_2$-CAE model—a CNN-based encoder-decoder model using the reconstruction error—was used as the baseline. Specifically, the performance of the proposed method was compared and evaluated based on the performance of the $l_2$–CAE model, which was introduced in \cite{bergmann2019mvtec}. In the research by \cite{bergmann2019mvtec}, the $l_2$-CAE model showed the best anomaly detection and localization performance compared with other models (SSIM-AE, GAN-based method, GMM-based method, etc.) on the MVTecAD dataset. Therefore, the $l_2$-CAE model, which recorded the highest performance in previous studies, was set as the baseline in this study. The $l_2$-CAE model of \cite{bergmann2019mvtec} adopts the CAE structure of \cite{DBLP:conf/visapp/BergmannLFSS19}. Hence, the $l_2$-CAE model was re-implemented in this study  by referring to these two papers, and performance was reported based on it.

The area under the ROC curve (AUROC), which has been used to verify the performance of anomaly detection and localization in several previous studies, was used as the indicator for performance evaluation \cite{ruff2018deep, bergmann2019mvtec, yi2020patch}. AUROC evaluates anomaly detection and localization performance based on the false positive rate (FPR) and true positive rate (TPR). This indicator has the advantage of not having to calculate the threshold using some of the test data because the threshold for abnormality needs not to be set deterministically.

\subsection{Implementation Details}
The encoder of AnoViT uses the same structure of the ViT model, and the model was pretrained on the ImageNet-21k dataset, the ViT weights finetuned on the ImageNet2012 dataset were used, and the model weights provided by the timm library were used. The weights were initialized with the model weights containing the information about the image representation, and image embeddings for anomaly detection were extracted through additional training. The patch size was set to 16, the embedding dimension was set to 768, and the number of heads was set to 8, as in the existing ViT-Base. The decoder comprised six blocks composed of transposed convolutional layer and ReLU activation function. To convert decoded image into the same size as the original image, an upsampling layer is added to the final layer of the decoder. 
The same image augmentation techniques were applied to the proposed method and the $l_2$-CAE model to compare their performance. The techniques used included random affine, vertical flip, horizontal flip, etc. Furthermore, Gaussian smoothing, which is a postprocessing technique applied in several previous studies on image anomaly detection \cite{defard2021padim, zavrtanik2021reconstruction, cohen2020sub} was applied to derive the final score map through dispersing the score for each pixel into a Gaussian distribution.

\section{Experimental Results}
\subsection{Anomaly Detection \& Localization Performance}
To evaluate the performance of the AnoViT with the ViT-based encoder-decoder structure, the anomaly detection and localization performance of the AnoViT was compared with that of the $l_2$-CAE model using the MNIST, CIFAR10, and MVTecAD datasets. First, the results in terms of anomaly detection using three types of data (MNIST, CIFAR10, and MVTecAD) confirmed that the proposed method exhibited better performance than the $l_2$-CAE model. 

\begin{table}[!ht]
\caption{Performance of anomaly detection methods on MNIST.}
\label{tab:det_mnist}
\centering 
\renewcommand{\arraystretch}{1.0}
\setlength{\tabcolsep}{4.6pt}
\begin{tabular}{llccc}
\toprule
\multirow{2}{*}{\textbf{Type}} & \multirow{2}{*}{\textbf{Category}} &  \multirow{2}{*}{$l_2$\textbf{-CAE}}   &  \textbf{Ours} &  \textbf{Relative}  \\ 
& & &  \textbf{(AnoViT)} & \textbf{Improvement} \\
\midrule
 \multirow{10}{*}{\shortstack[l]{Class}} 
    & 0  & 0.938 & \uline{\textbf{0.980}} & \uline{\textbf{4.478\%}} \\ 
    & 1  & 0.994 & 0.986 & -0.805\% \\
    & 2  & 0.853 & \uline{\textbf{0.911}} & \uline{\textbf{6.800\%}} \\ 
    & 3  & 0.880 & 0.838 & -4.772\% \\  
    & 4  & 0.878 & \uline{\textbf{0.913}} & \uline{\textbf{4.000\%}} \\  
    & 5  & 0.899 & \uline{\textbf{0.924}} & \uline{\textbf{2.780\%}} \\  
    & 6  & 0.949 & \uline{\textbf{0.975}} & \uline{\textbf{2.740\%}} \\ 
    & 7  & 0.925 & 0.914 & -1.189\% \\  
    & 8  & 0.834 & \uline{\textbf{0.838}} & \uline{\textbf{0.480\%}} \\ 
    & 9  & 0.926 & \uline{\textbf{0.958}} & \uline{\textbf{3.456\%}} \\ \hline
 
 \multirow{1}{*}{\shortstack[l]{mean}} &  -  & 0.908 & \uline{\textbf{0.924}} & \uline{\textbf{1.774\%}}  \\
 \multirow{1}{*}{\shortstack[l]{std}} & - & 0.044 & 0.049 & -  \\
\bottomrule
\end{tabular}
\end{table}

Table \ref{tab:det_mnist} presents the anomaly detection performance on the MNIST dataset. It shows that the performance improved relative to the baseline in 7 out of 10 classes, and the average AUROC increased by 1.774\% compared with the $l_2$-CAE model.
Table \ref{tab:det_cifar10} shows the results of the experiment conducted with the CIFAR10 dataset. The proposed ViT-based proposed method achieved improved results compared with the $l_2$-CAE model in 7 out of 10 classes. 

\begin{table}[h!]
\caption{Performance of anomaly detection methods on CIFAR10.}
\label{tab:det_cifar10}
\centering 
\renewcommand{\arraystretch}{1.0}
\setlength{\tabcolsep}{4pt}
\begin{tabular}{llccc}
\toprule
\multirow{2}{*}{\textbf{Type}} & \multirow{2}{*}{\textbf{Category}} &  \multirow{2}{*}{$l_2$\textbf{-CAE}}   &  \textbf{Ours} &  \textbf{Relative}  \\ 
& & &  \textbf{(AnoViT)} & \textbf{Improvement} \\
\midrule
 \multirow{10}{*}{\shortstack[l]{Class}} 
  & Airplane    & 0.597 & \uline{\textbf{0.664}} & \uline{\textbf{11.222\%}} \\ 
  & Automobile  & 0.525 & 0.464 & -11.619\% \\ 
  & Bird        & 0.585 & \uline{\textbf{0.639}} & \uline{\textbf{9.231\%}} \\ 
  & Cat         & 0.525 & \uline{\textbf{0.550}} & \uline{\textbf{4.762\%}} \\ 
  & Deer        & 0.644 & \uline{\textbf{0.743}} & \uline{\textbf{15.373\%}} \\ 
  & Dog         & 0.547 & 0.542 & -0.914\% \\ 
  & Frog        & 0.638 & \uline{\textbf{0.700}} & \uline{\textbf{9.718\%}} \\ 
  & Horse       & 0.428 & \uline{\textbf{0.509}} & \uline{\textbf{18.925\%}} \\ 
  & Ship        & 0.675 & \uline{\textbf{0.697}} & \uline{\textbf{3.259\%}} \\  
  & Truck       & 0.526 & 0.502 & -4.563\% \\ 
 \hline
 \multirow{1}{*}{\shortstack[l]{mean}} & -  & 0.569 & \uline{\textbf{0.601}} & \uline{\textbf{5.624\%}}  \\
 \multirow{1}{*}{\shortstack[l]{std}} & - & 0.069 & 0.093 & -  \\
\bottomrule
\end{tabular}
\end{table}

\begin{table}[b!]
\caption{Performance of anomaly detection methods on MVTecAD.}
\label{tab:det_mvtec}
\centering 
\renewcommand{\arraystretch}{1.0}
\setlength{\tabcolsep}{4pt}
\begin{tabular}{llccc}
\toprule
\multirow{2}{*}{\textbf{Type}} & \multirow{2}{*}{\textbf{Category}} &  \multirow{2}{*}{$l_2$\textbf{-CAE}}   &  \textbf{Ours} &  \textbf{Relative}  \\ 
& & &  \textbf{(AnoViT)} & \textbf{Improvement} \\
\midrule
 \multirow{10}{*}{\shortstack[l]{Object}} 
    & Bottle     & 0.77  & \uline{\textbf{0.83}} & \uline{\textbf{7.79\%}} \\ 
    & Cable      & 0.66	 & \uline{\textbf{0.74}} & \uline{\textbf{12.12\%}}  \\   
    & Capsule    & 0.67  & \uline{\textbf{0.73}} & \uline{\textbf{8.96\%}} \\ 
    & Hazelnut   & 0.88  & 0.88 & 0.00\% \\  
    & Metal Nut  & 0.42 & \uline{\textbf{0.86}} & \uline{\textbf{104.76\%}} \\  
    & Pill       & 0.68 & \uline{\textbf{0.72}} & \uline{\textbf{5.88\%}} \\ 
    & Screw      & 1.00 & 1.00 & 0.00\% \\ 
    & Toothbrush & 0.41 & \uline{\textbf{0.74}} & \uline{\textbf{80.49\%}} \\  
    & Transistor & 0.88 & 0.83 & -5.68\% \\
    & Zipper     & 0.71 & \uline{\textbf{0.73}} & \uline{\textbf{2.81\%}} \\ \hline
  
 \multirow{5}{*}{\shortstack[l]{Texture}} 
    & Carpet     & 0.54 & 0.50 & -7.41\% \\ 
    & Grid       & 0.78 & 0.52 & -33.33\% \\  
    & Leather    & 0.77 & \uline{\textbf{0.85}} & \uline{\textbf{10.39\%}} \\ 
    & Tile       & 0.85 & \uline{\textbf{0.89}} & \uline{\textbf{4.71\%}} \\ 
    & Wood       & 0.98 & 0.95 & -3.06\% \\ \hline

 \multirow{1}{*}{\shortstack[l]{mean}} & -  & 0.73 & \uline{\textbf{0.78}} & \uline{\textbf{6.85\%}} \\ 
 \multirow{1}{*}{\shortstack[l]{std}} & - & 0.18 & 0.14 & - \\ 
\bottomrule
\end{tabular}
\end{table}

Moreover, it may also be noted that the proposed method detected anomalies relatively better than the $l_2$-CAE model by over 5.624\% on average. In addition, anomaly detection was performed using the MVTecAD dataset. The results on MVTecAD dataset in Table \ref{tab:det_mvtec} show that the proposed model exhibited better performance than the $l_2$-CAE model in 9 out of 15 categories. Also, it may be observed from Table \ref{tab:det_mvtec} that the proposed method performed anomaly detection well in various types (texture, object) of product images.

\begin{table}[!b]
\caption{Performance of anomaly localization methods on MVTecAD.}
\label{tab:loc_mvtec}
\centering 
\renewcommand{\arraystretch}{1.0}
\setlength{\tabcolsep}{4pt}
\begin{tabular}{llccc}
\toprule
\multirow{2}{*}{\textbf{Type}} & \multirow{2}{*}{\textbf{Category}} &  \multirow{2}{*}{$l_2$\textbf{-CAE}}   &  \textbf{Ours} &  \textbf{Relative}  \\ 
& & &  \textbf{(AnoViT)} & \textbf{Improvement} \\
\midrule
 \multirow{10}{*}{\shortstack[l]{Object}} 
    & Bottle     & 0.83 & \uline{\textbf{0.86}} & \uline{\textbf{3.61\%}} \\
    & Cable      & 0.85 & \uline{\textbf{0.89}} & \uline{\textbf{4.71\%}} \\ 
    & Capsule    & 0.91 & 0.91 & 0.00\% \\  
    & Hazelnut   & 0.95 & 0.94 & -1.05\% \\  
    & Metal Nut  & 0.86 & \uline{\textbf{0.88}} & \uline{\textbf{2.33\%}} \\ 
    & Pill       & 0.85 & \uline{\textbf{0.86}} & \uline{\textbf{1.18\%}} \\  
    & Screw      & 0.92 & 0.92 & 0.00\% \\
    & Toothbrush & 0.85 & \uline{\textbf{0.90}} & \uline{\textbf{5.88\%}} \\ 
    & Transistor & 0.77 & \uline{\textbf{0.80}} & \uline{\textbf{3.90\%}} \\ 
    & Zipper     & 0.74 & \uline{\textbf{0.76}} & \uline{\textbf{2.70\%}} \\ \hline
  
 \multirow{5}{*}{\shortstack[l]{Texture}} 
    & Carpet     & 0.71 & 0.65 & -8.45\% \\ 
    & Grid       & 0.67 & \uline{\textbf{0.83}} & \uline{\textbf{23.88\%}} \\  
    & Leather    & 0.82 & \uline{\textbf{0.89}} & \uline{\textbf{8.54\%}} \\
    & Tile       & 0.59 & 0.57 & -3.40\% \\
    & Wood       & 0.82 & \uline{\textbf{0.85}} & \uline{\textbf{3.66\%}} \\ \hline
  
 \multirow{1}{*}{\shortstack[l]{mean}} & -  & 0.81 & \uline{\textbf{0.83}} & \uline{\textbf{2.47\%}} \\ 
 \multirow{1}{*}{\shortstack[l]{std}} & - & 0.10 & 0.10 & - \\ 
\bottomrule
\end{tabular}
\end{table}

\begin{figure*}[t!] 
\begin{center}
\includegraphics[width=\linewidth]{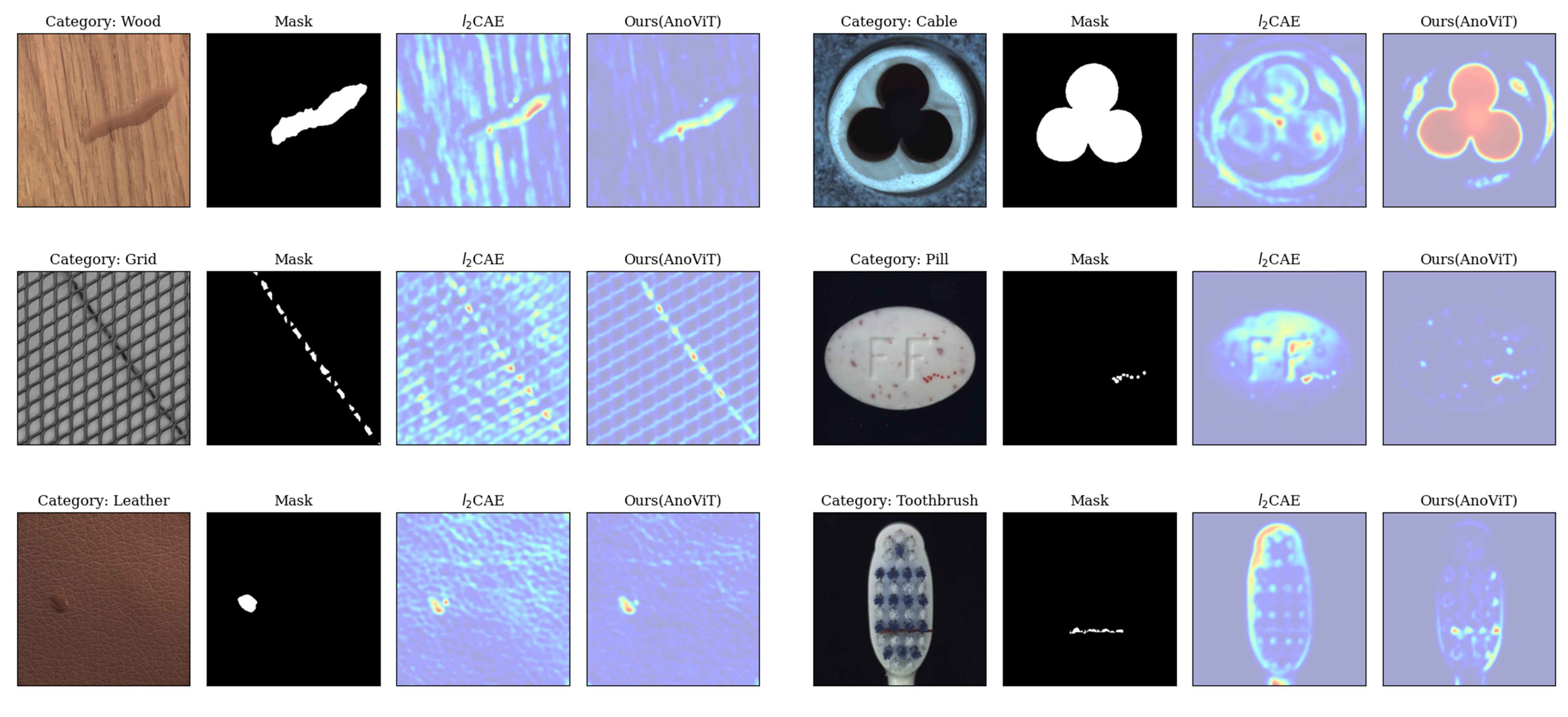}
\end{center}
\caption{Qualitative result of anomaly localization in MVTecAD.}
\label{fig3}
\end{figure*} 

Moreover, Table \ref{tab:loc_mvtec} shows the results of evaluating the anomaly localization performance of the proposed method using the MVTecAD dataset. The performance results of the proposed method were compared with that of the $l_2$-CAE model, and the anomaly localization performance improved in 10 of 15 categories. In two categories, the performance of the proposed method was the same as that of the $l_2$-CAE model. The localization performance improved the average AUROC by 2.47\% compared to $l_2$-CAE model, regardless of the product type in the image.

In the experiment where ViT was used as the encoder model instead of CNN under the same conditions, such as training and data preprocessing, the results show that the anomaly detection and localization performance improved. This outcome can be attributed to the fact that the global information, as well as the local information, was additionally reflected in the patch embedding through multiple self-attention operations, which comprise the ViT.

\subsection{Qualitative Analysis}

The results of the localization task were qualitatively evaluated by visualizing the anomaly score map of the proposed method on the MVTecAD dataset. Fig. \ref{fig3} shows the localization results of the texture and object types of the MVTecAD dataset. The three images on the left in Fig. \ref{fig3} are texture-type products, and the three on the right show the results for the object-type products. For each product, the first column shows the abnormal product image, the second column shows the ground-truth mask, the third column is the score map of the $l_2$-CAE model, and the last column is the score map of the proposed method. The figure below shows low scores in blue and high scores in red. That is, the closer the area lies to the red color, the higher the abnormal score threshold, which indicates a high possibility of being an anomalous area.

Fig. \ref{fig3} confirms qualitatively that the proposed method detected anomalous areas well for all types of images, regardless of the size and shape of the abnormal area. Even when a large anomalous area was included, such as cable products, and when the anomalous area was very small, such as a toothbrush or pill, the ViT-based encoder-decoder model accurately predicted the location of the abnormal areas compared with the $l_2$-CAE model. In particular, in the score map of the wood and grid product images, the $l_2$-CAE model derived a high abnormal score for the abnormal areas as well as the pixels in the normal region. Hence, an area wider than the ground- truth mask was determined to be abnormal. By contrast, the abnormal scores of pixels in the area outside the anomalous area were low for the proposed method. Hence, the proposed method found abnormal areas that were very similar to the ground-truth mask.

\section{Conclusion}
Image anomaly detection and localization have been used to detect anomalies such as defects and abnormal behaviors in image data in the fields of manufacturing, medical care, and intelligent information. Image anomaly detection and localization have also been used to improve work efficiency and help experts make accurate decisions. In image anomaly detection using encoder-decoder models, it is essential to derive embeddings containing rich normal information because a reconstruction error is calculated using an image reconstructed from a representation containing the normal information.

Therefore, we have proposed a ViT-based encoder-decoder designed to perform both anomaly detection and localization. Abnormal images and areas are detected using the reconstruction error calculated from the proposed method. In contrast to CNN-based encoder-decoder models that only reflect local information in an image, the ViT encoder extracts a representation that contains the normal information in detail, including the relationship between the image patches. In addition, the patch embeddings derived from the encoder are used to construct a three-dimensional feature map and sent to the decoder to preserve the spatial information well. The experimental result confirmed that the performance of the proposed method was better than the CNN-based $l_2$-CAE model on both quantitative and qualitative evaluations. 

To demonstrate that anomaly detection and localization performance were improved by extracting a normal representation with a self-attention operation-based encoder rather than a model using convolution operations among encoder-decoder-based methods, a CNN-based $l_2$-CAE model was selected for comparison in this study. In future research, further performance evaluation should be performed between the proposed method and several encoder-decoder-based models, in addition to the $l_2$-CAE model. Furthermore, if a subsequent study demonstrates that the anomaly detection and localization performance is improved when DeiT\cite{touvron2021training} or CrossViT\cite{chen2021crossvit}, which were developed from the basic ViT model, is used as the encoder, this result is expected to lead to further active investigation of vision transformer-type models for anomaly detection. 

\bibliography{main}
\bibliographystyle{icml2021}
\clearpage

\end{document}